\documentclass[a4paper,conference]{IEEEtran}
\IEEEoverridecommandlockouts
\usepackage{cite}
\usepackage{amsmath,amssymb,amsfonts}
\usepackage{algorithmic}
\usepackage{graphicx}
\usepackage{textcomp}
\usepackage{xcolor}
\usepackage{booktabs}
\usepackage{amssymb}
\usepackage{float}
\usepackage{listings}
\usepackage[colorlinks]{hyperref}
\def\BibTeX{{\rm B\kern-.05em{\sc i\kern-.025em b}\kern-.08em
    T\kern-.1667em\lower.7ex\hbox{E}\kern-.125emX}}

\ifCLASSOPTIONcompsoc
\usepackage[caption=false,font=normalsize,labelfon
t=sf,textfont=sf]{subfig}
\else
\usepackage[caption=false,font=footnotesize]{subfi
g}
\fi

\begin{document}

\title{SAILenv: Learning in Virtual Visual \\ Environments Made Simple}

\author{\IEEEauthorblockN{\hskip 16mm Enrico Meloni$^{1,2}$, Luca Pasqualini$^1$,}
\IEEEauthorblockA{\textit{$^1$Dept. of Information Engineering and Mathematics} \\
\textit{University of Siena}\\
Siena, Italy}
\and
\IEEEauthorblockN{\hskip -14mm Matteo Tiezzi$^1$, Marco Gori$^{1,3}$,}
\IEEEauthorblockA{\textit{$^2$Dept. of Information Engineering} \\
\textit{University of Florence}\\
Florence, Italy}
\and
\IEEEauthorblockN{\hskip -30mm Stefano Melacci$^1$}
\IEEEauthorblockA{\textit{$^3$MAASAI} \\
\textit{Universit\`{e} C\^{o}te d'Azur}\\
Nice, France \\
\hskip -14cm \{meloni,pasqualini,mtiezzi,marco,mela\}@diism.unisi.it}
}

\maketitle

\begin{abstract}
Recently, researchers in Machine Learning algorithms, Computer Vision scientists, engineers and others, showed a growing interest in 3D simulators as a mean to artificially create experimental settings that are very close to those in the real world. However, most of the existing platforms to interface algorithms with 3D environments are often designed to setup navigation-related experiments, to study physical interactions, or to handle ad-hoc cases that are not thought to be customized, sometimes lacking a strong photorealistic appearance and an easy-to-use software interface. In this paper, we present a novel platform, SAILenv, that is specifically designed to be simple and customizable, and that allows researchers to experiment visual recognition in virtual 3D scenes. A few lines of code are needed to interface every algorithm with the virtual world, and non-3D-graphics experts can easily customize the 3D environment itself, exploiting a collection of photorealistic objects. Our framework yields pixel-level semantic and instance labeling, depth, and, to the best of our knowledge, it is the only one that provides motion-related information directly inherited from the 3D engine. The client-server communication operates at a low level, avoiding the overhead of HTTP-based data exchanges.
We perform experiments using a state-of-the-art object detector trained on real-world images, showing that it is able to recognize the photorealistic 3D objects of our environment. The computational burden of the optical flow compares favourably with the estimation performed using modern GPU-based convolutional networks or more classic implementations. We believe that the scientific community will benefit from the easiness and high-quality of our framework to evaluate newly proposed algorithms in their own customized realistic conditions.

\end{abstract}


\section{Introduction}
Developing Machine Learning algorithms to solve a target task usually follows a well-established offline process in which data are collected from the real-world operational environment and then used to learn the models parameters and to evaluate the quality of the trained model. 
Recently, the scientific literature started to focus on the precious role of realistic 3D virtual environments that simulate the real-world setting, allowing researchers to perform a variety of controlled tests that would be very costly if directly performed in the real-world \cite{habitat19iccv}. Models learned in the simulated environment might need to be adjusted to compensate the differences between the simulated and the real world, thus high-quality virtual environments are commonly needed to reduce the compensation effort \cite{ai2thor,habitat19iccv}.

If we depart from the case of the most popular benchmarks shared by the scientific community, such as the ones aimed at showing the quality of visual navigation algorithms \cite{zhu2017target,gupta2017cognitive}, visual QA \cite{gordon2018iqa}, and others \cite{chaplot2018gated,beattie2016deepmind}, each research project has its own characteristic features, and it actually requires to design the 3D environment that correctly resembles the target working conditions. Moreover, the way a virtual agent will exploit the information coming from the virtual world, and how it will react to it, need to be designed coherently with the target setting. This clearly suggests that there is the need of providing flexible and easy-to-use tools to encourage the use of virtual environments and to favour the development of those research activities that exploit them. Another important consideration to remark is that not all researchers have robust skills in creating 3D scenes, and this aspect might discourage the use of virtual environments.

With the aim of providing a simple and customizable platform to perform visual recognition experiments in 3D virtual environments, in this paper we present \textit{SAILenv}, the Siena Artificial Intelligence Laboratory\footnote{SAILab, \url{https://sailab.diism.unisi.it}.} environment, that can be freely downloaded following the instructions at \url{http://sailab.diism.unisi.it/sailenv}, where an extended tutorial is also included. SAILenv is based on  Unity, a popular game engine that supports several platforms, developed by Unity Technologies\footnote{See \url{https://unity.com} for further details.}, and that includes advanced 3D modeling and state-of-the art quality real-time rendering. SAILenv provides a Python interface that, with a few lines of code, allows every algorithm to get data from the 3D world handled by Unity. To the best of our knowledge, SAILenv is the only platform that yields real-time motion-related information inherited from the 3D engine (thus being extremely accurate), and not computed afterwards from multiple 2D observations, as commonly done in optical flow algorithms \cite{flownetlite,farneback2003two}. This is an important feature when developing Computer Vision algorithms that benefit from the motion field. Moreover, differently from other popular platforms \cite{ai2thor}, the data in SAILenv are transferred without relying on any higher-level communication protocols (such as HTTP), thus without introducing further overhead in the transmission. 
Unity provides a powerful editor to customize the 3D environment, that, however, might discourage researchers when starting from scratch. To overcome this issue, SAILenv includes a ready-to-use Unity project with a set of objects that were initially taken from AI2-THOR 2.1.0 \cite{ai2thor}, and that we augmented adding realistic textures and lighting effects by means of state-of-the art texturing software, making them strongly photorealistic. Assembling a scene using the provided Unity assets can be done in a few steps, making the virtual environment ready to be queried by the Python code.


\begin{figure*}[!t]
\centering
\subfloat[Bed]{\includegraphics[width=0.15\textwidth]{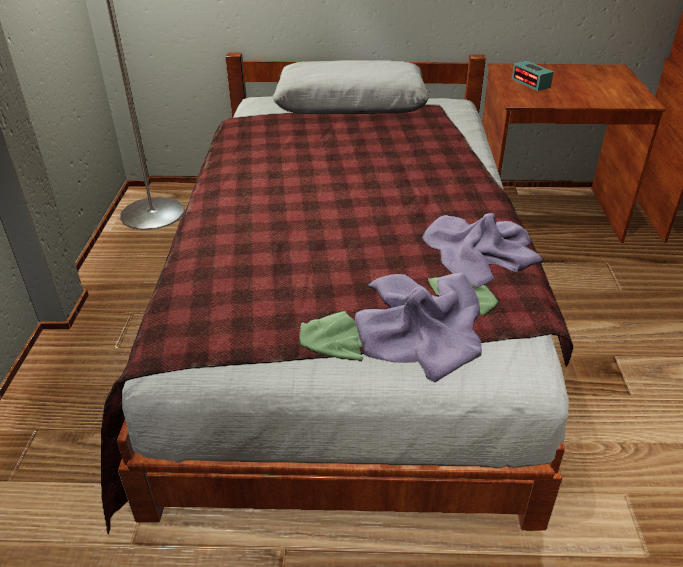}
\label{fig:bed}}
\hfil
\subfloat[Cabinet]{\includegraphics[width=0.15\textwidth]{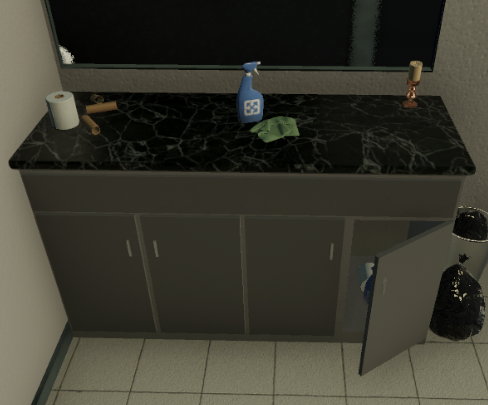}
\label{fig:cabinet}}
\hfil
\subfloat[Couch]{\includegraphics[width=0.15\textwidth]{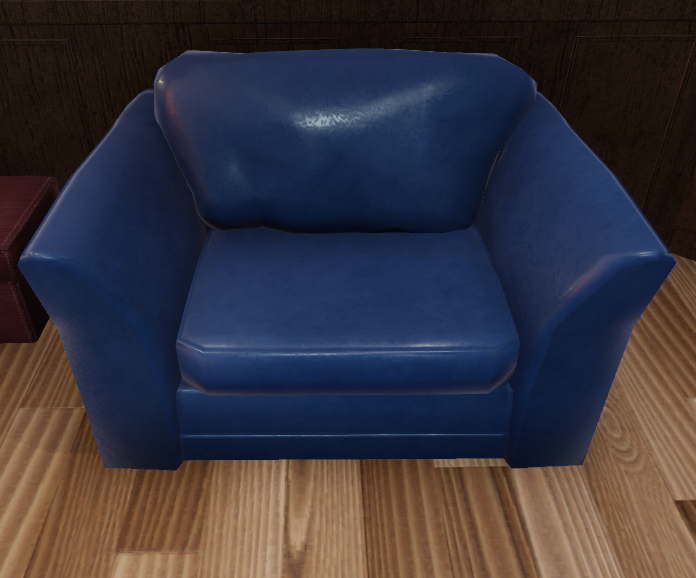}
\label{fig:couch}}
\hfil
\subfloat[Laptop]{\includegraphics[width=0.15\textwidth]{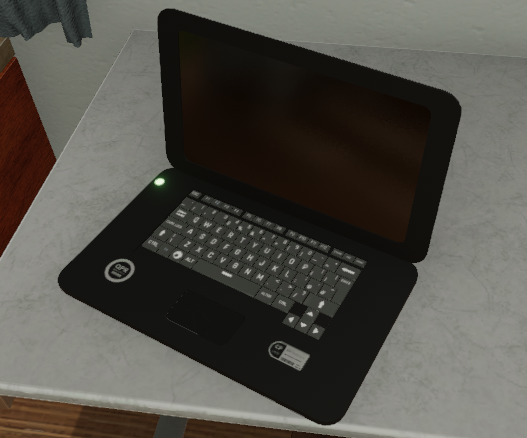}
\label{fig:laptop}}
\hfil
\subfloat[Table]{\includegraphics[width=0.15\textwidth]{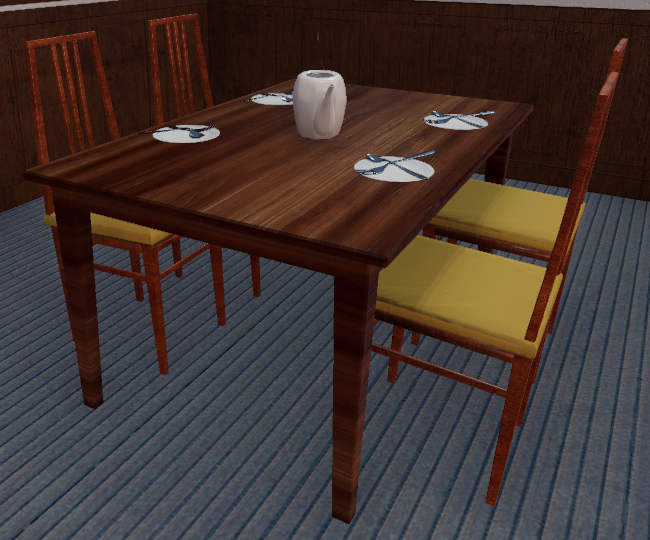}
\label{fig:table}}
\caption{Samples of objects available in SAILenv.}
\label{fig:objects}
\end{figure*}

We provide experimental evidence on the quality of the scenes/objects included in SAILenv, showing that a state-of-the art neural model \cite{matterport_maskrcnn_2017} trained on real-world data can easily recognize the SAILenv objects. We also measure and compare the speed in computing motion features, showing that SAILenv leads to smaller running times than the ones of popular optical flow estimators (including the ones based on convolutional neural networks \cite{flownetlite}), being intrinsically more accurate.

This paper is organized as follows. Section~\ref{sec:related} describes related 3D frameworks, emphasizing the differences with SAILenv. The structure of SAILenv is discussed in Section~\ref{sec:architecture}, while usage examples are in Section~\ref{sec:using}. Section~\ref{sec:exp} is about our experimental activity while Section~\ref{sec:conclusions} concludes the paper.

\section{Related Platforms}
\label{sec:related}
Several environments and simulators have been developed by the scientific community in the last few years. 
Some simulators are not photorealistic, or they are specifically designed to handle specific tasks. Some examples are DeepMind Lab \cite{beattie2016deepmind}, UETorch \cite{lerer2016learning}, Scene \cite{handa2016scenenet}.
Amongst the virtual environments with visual-realistic appearance we mostly focus on the recent AI2-THOR \cite{ai2thor} and Habitat \cite{habitat19iccv}. Other existing frameworks are Home \cite{brodeur2017home}, Chalet \cite{yan2018chalet}, Gibson \cite{xia2018gibson}, SceneNet RGBD \cite{McCormac_2017_ICCV}.
These environments are used to study embodied agents \cite{xia2018gibson}, to instantiate tasks that are about visual navigation with reinforcement learning \cite{zhu2017target,gupta2017cognitive}, interactive VQA \cite{gordon2018iqa}, task-oriented language grounding \cite{chaplot2018gated} or vision-and-language navigation \cite{wang2018look}.

SAILenv, coherently with what is commonly done in related platforms, captures RGB representations with or without depth information, acquired from the agent camera position and orientation.
Similarly to what we propose, also AI2-THOR \cite{ai2thor} is based on the Unity engine, but it focuses on the interaction with the environment, so that actions can be attached to objects. Differently, SAILenv focuses on visual recognition, and it simplifies the addiction of new semantic categories to objects, an operation that does not require knowledge of the code structure, and that can be done through the Unity GUI. Moreover, the client-server architecture of AI2-THOR is based on HTTP communication between Unity and the Python API, where the 3D engine acts as a client while the server is implemented on the Python side of the architecture. SAILenv, as we will describe in Section~\ref{sec:architecture}, implements a more natural organization in which the virtual world is a server to which a Python client gets connected to retrieve data that will be processed by the target algorithm. Habitat \cite{habitat19iccv} is mostly focused in allowing the access to different 3D datasets (such as \cite{song2017semantic,chang2017matterport3d}) by a uniform interface, and it includes its own (fast) simulation engine. While in principle the direct customization or the creation of 3D environments is possible, it is not straightforward. Differently, SAILenv is built around the Unity engine, that is a very popular and multi-platform software solution. In Table~\ref{tab:comparison} we summarize a comparison of the main features of SAILenv with some of the aforementioned frameworks. Notice that only SAILenv includes optical flow and lightweight communication over the network, and that, differently from AI2-THOR and Habitat, it can also run on a Windows machine.

\begin{table}[t]
\caption{Comparison of the main features of SAILenv with other popular platforms. \textit{LightNet} refers to lightweight communication over the network (n.a. means network communication is not directly provided).}
\label{tab:comparisons}
\centering
\begin{tabular}{l@{$\ $}|@{$\ $}c@{$\ $}|@{$\ $}c@{$\ $}|@{$\ $}c@{$\ $}|@{$\ $}c@{$\ $}|@{$\ $}c@{$\ $}}
\toprule
\scriptsize \textsc{Platform}	& \scriptsize \textsc{Photoreal} & \scriptsize \textsc{Depth} & \scriptsize \textsc{OptFlow} & \scriptsize \textsc{LightNet} & \scriptsize \textsc{OS} \\
\midrule
DeepMind Lab \cite{beattie2016deepmind} & & \checkmark & & n.a. & Unix\\
Habitat \cite{habitat19iccv} & \checkmark & \checkmark & & n.a. & Unix\\ 
AI2-THOR \cite{ai2thor} & \checkmark & \checkmark & & & Unix \\
\textbf{SAILenv} & \checkmark & \checkmark & \checkmark & \checkmark & Win+Unix \\
\bottomrule
\end{tabular}

\label{tab:comparison}
\end{table}

\section{Architecture and Main Features}
\label{sec:architecture}

\begin{figure}[H]
\begin{center}
\includegraphics[width=0.85\linewidth]{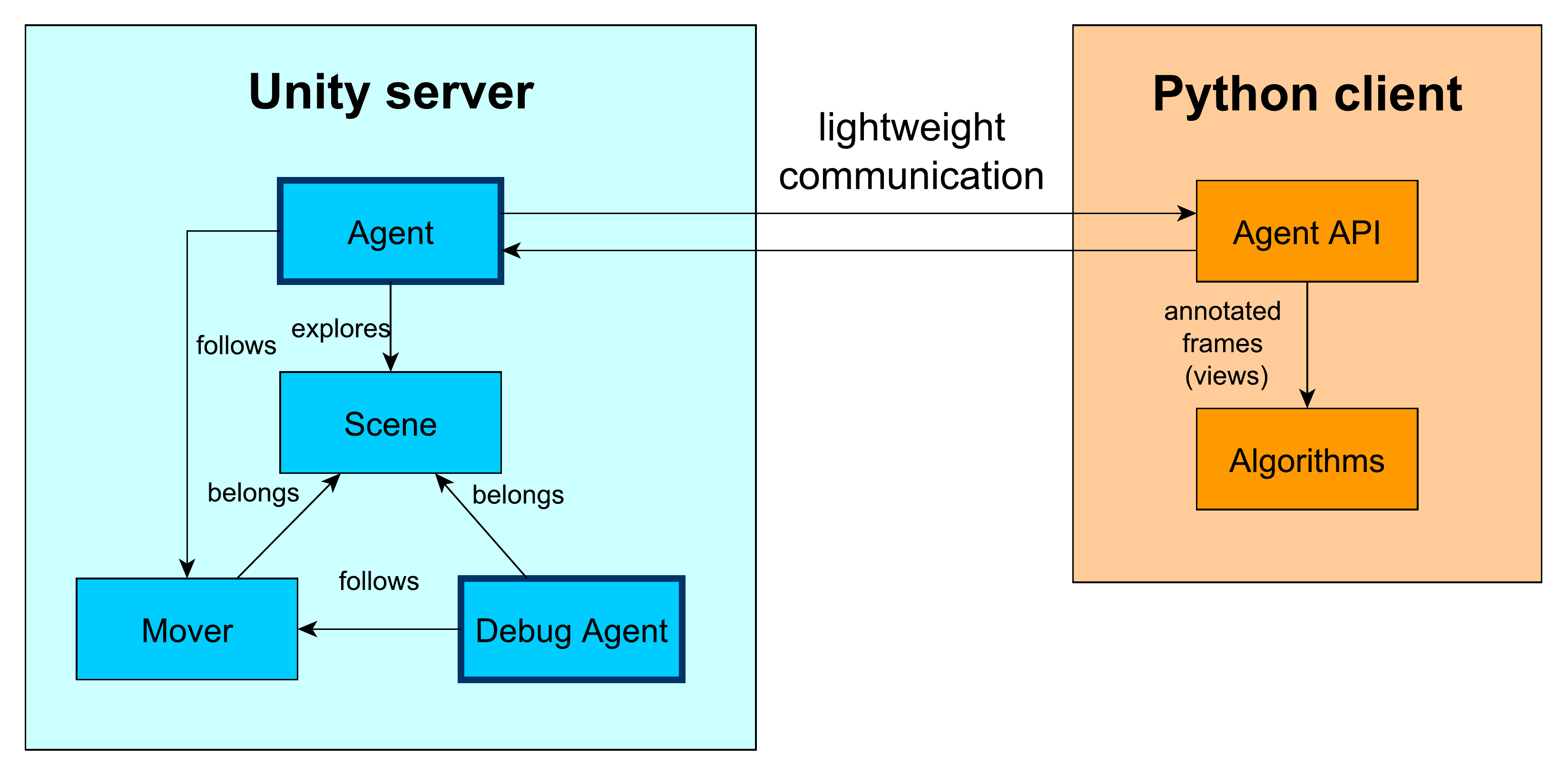}
\caption{Organization of the SAILenv architecture.}
\label{fig:system_architecture}
\end{center}
\end{figure}
SAILenv is organized following a client-server architecture that naturally implements the idea of having a virtual \textit{scene} (server) and an \textit{agent} that explores it (manipulated by the client).
The agent position and orientation are changed by means of the client commands, implemented in a simple Python API, also referred to as \textit{agent API}. Whenever the client queries the environment for information, the agent returns a number of \textit{views} that capture different properties of what the agent is observing, i.e., annotated frames. The view data can then be processed by the target  Computer Vision algorithm, or fed to existing Machine Learning frameworks, libraries or other software. The overall architecture of SAILenv is summarized in Fig.~\ref{fig:system_architecture}.

\begin{figure*}[!t]
\centering
\subfloat[Category View]{\includegraphics[width=0.2\textwidth]{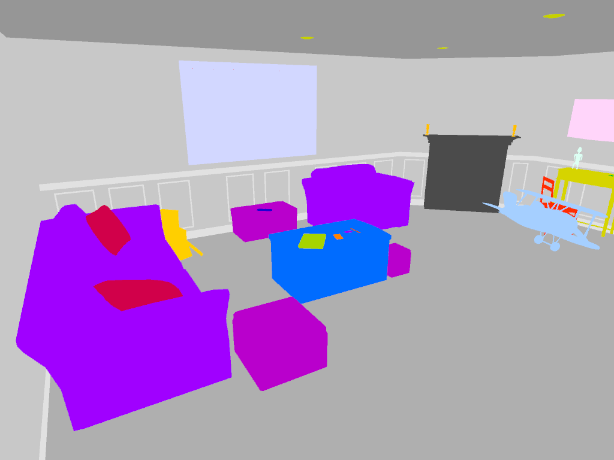}
\label{fig:category}}
\hfil
\subfloat[Instance View]{\includegraphics[width=0.2\textwidth]{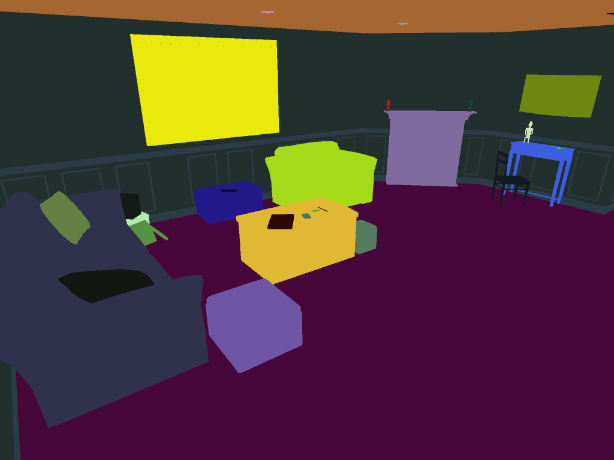}
\label{fig:instance}}
\hfil
\subfloat[Depth View]{\includegraphics[width=0.2\textwidth]{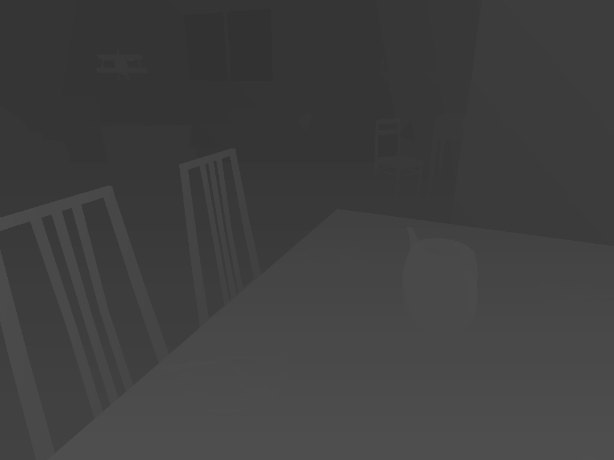}
\label{fig:depth}}
\caption{Pixel-wise annotations yielded by SAILenv (Optical flow is reported in Fig.~\ref{fig:optical_flow})}
\label{fig:annotations}
\end{figure*}

We implemented the server within the Unity framework, creating an ad-hoc \textit{Unity server} that waits for client requests. The Unity server is more than just a network interface layer, since it is a computational module that is responsible of creating the virtual environment, managing the physics simulation and the real-time rendering, fully exploiting the Unity infrastructure. It also generates and packs the data attached to the views requested by the client. Thanks to the power of the Unity environment, the physics simulation runs at real-time speed on most of nowadays servers or laptops. 
The Unity instance that is running on the server allows the server screen to show what the agent camera is currently capturing (for debug purposes) and, as it will become clearer in the rest of this section, it supports some special server-side interactions.
The \textit{Python client} includes a lightweight cross-platform API with a small dependency tree, and it is in fact a tiny interface that exposes high-level commands over the virtual environment, such as creating a new agent, moving the agent or obtaining views of the current state of the environment through the 
``eyes'' of the agent. In particular, SAILenv yields  the classic RGB view, depth information, optical flow, semantic segmentation, instance labeling, as shown in Fig.~\ref{fig:annotations} and Fig.~\ref{fig:optical_flow}. Each view includes information for all the pixels acquired by the camera. The RGB view is straightforwardly implemented representing each pixel (in each channel) with the classic 8-bit encoding, while depth information is inherited by the Unity engine, through a gray-scale texture representing the distance of the observed rendered objects from the camera view. In the depth example of Fig.~\ref{fig:annotations}c, lighter pixels indicate elements that are closer to the agent.
Each pixel is fully annotated with a category identifier (semantic labeling) and an instance identifier, that are encoded in the category and instance views, respectively, as show in Fig.~\ref{fig:annotations}a-b. While the instance identifier is implicitly given by the unique Unity identifier of each object in the scene, categories can be added or edited using the Unity editor GUI, without any code-level operations. In particular, categories are represented as Unity objects, and they can be attached to every other object by a drag-and-drop operation. SAILenv also includes a ``category holder'' to organize sets of categories and to allow the user to quickly add them to the current scene.
We post-pone the description of the view associated to optical flow to Section~\ref{sec:flow}.
Whenever all the views requested by the client are ready, they are sent over a network communication as raw bytes or GZip encoded bytes. This makes the communication efficient, avoiding possible overhead caused by HTTP-based communication or other higher-level data-packing and transmission protocols, as we will describe in Section~\ref{sec:socket}. 


SAILenv allows users to have access to a collection of photo-realistic objects to simplify the development of novel scenes, whose visual quality has been enhanced using modern effects and shaders, that will be discussed in Section~\ref{sec:photo}.  
Movement dynamics can be attached to the objects of the scene with a few operations, eventually using the movement templates of Section~\ref{sec:moveobj}, while the way the agent movement is controlled is due to the \textit{mover} element of  Fig.~\ref{fig:system_architecture}, that will be the subject of Section~\ref{sec:mover}.

\subsection{Realistic Optical Flow}
\label{sec:flow}
\begin{figure*}[!t]
\centering
\subfloat[Camera Motion Optical Flow]{\includegraphics[width=0.2\textwidth]{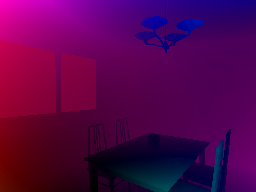}
\label{fig:flow1}}
\hfil
\subfloat[Rotating Object Optical Flow]{\includegraphics[width=0.2\textwidth]{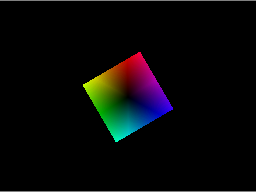}
\label{fig:flow2}}
\hfil
\subfloat[LiteFlowNet Optical Flow]{\includegraphics[width=0.2\textwidth]{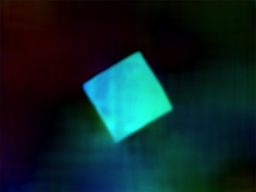}
\label{fig:flow_flownet}}
\hfil
\subfloat[OpenCV Optical Flow]{\includegraphics[width=0.2\textwidth]{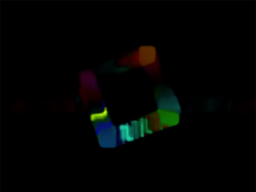}
\label{fig:flow_cv}}
\caption{Optical flow yielded by SAILenv. In \autoref{fig:flow1} we see the optical flow given by the agent motion. In \autoref{fig:flow2} we see the optical flow given by the objects motion, while the agent stands still. In \autoref{fig:flow_flownet} we show how LiteFlowNet \cite{flownetlite} estimates the optical flow in the same conditions, while in \autoref{fig:flow_cv} we show the estimation of the OpenCV implementation of the Farneback Optical Flow.}
\label{fig:optical_flow}
\end{figure*}

SAILenv also yields highly precise and dense motion information about the environment. Differently from what is done by the most common optical flow algorithms, the SAILenv optical flow is not due to an estimation obtained by observing consecutive frames, and it is fully computed by the physics engine of Unity. 
Unity has access to the information about the motion of the objects in the scene and the agent viewpoint, and it uses them to drive the simulation of physics of the environment. SAILenv inherits such information and  adapts it to generate a view that includes the motion vectors for all the pixels of the frame. In detail, such view is a $H \times W \times 2$ tensor of single-precision floating point numbers, being $H$ and $W$ the height and width of the frame, respectively. For each pixel, a pair of floats describes the velocity of the pixel (pixels per second). An example of such motion field, represented in the HSV color space, is reported in Fig.~\ref{fig:optical_flow}(a).\footnote{Cartesian coordinates $(x, y)$ are first converted into polar coordinates $(\alpha,  \theta)$ where $\alpha$ is the magnitude and $\theta$ is the phase. Then, then we set $H = \theta$, $S = 1$, and $V = \alpha$.}

In Fig.~\ref{fig:optical_flow}b-d, we report three examples of the optical flow computed in a scene populated solely by a rotating cube. The cube has no special textures, and it has a uniform color. This clearly makes it hard to estimate the pixel-level motion using classic algorithms, while SAILenv can correctly capture the rotation of the cube (Fig.~\ref{fig:optical_flow}b). Widely used implementations, such as the Farneback algorithm implemented in OpenCV\footnote{\url{https://opencv.org/}}, or modern approaches based on convolutional neural networks \cite{flownetlite} fail in correctly capturing the motion, as noticeable in  Fig.~\ref{fig:optical_flow}c-d.
Despite its very high precision, the optical flow has almost null computational burden on the Unity server. Of course, some overhead is due to data normalization operations, that is still negligible with respect to what is commonly needed to infer motion from pairs of static frames, as we will experiment in Section~\ref{sec:exp}.

\subsection{Lightweight Network Transmission}
\label{sec:socket}
The Unity environment awaits for incoming connections on a target port. By the time the Python client connects to the socket, calling the apposite API, the generation of an agent inside the 3D environment is triggered. The agent is associated with a worker background thread which listens and replies to further requests. 
Even if the worker thread is implemented within the Unity framework, it is not synchronous with respect to the physics engine, in order to avoid network-related slowdowns or communication issues to lock the simulation engine. Instead of using any higher-level protocols to serialize the data and send it through the network, SAILenv either sends the raw data or GZipped data (customizable), reducing at its minimum the communication overhead.
This is different from other popular solutions that rely on other software stacks to handle the data management \cite{ai2thor}.

\subsection{Photo-realistic Objects and Scenes}
\label{sec:photo}
Relying on the Unity engine to handle the virtual environments allows SAILenv to exploit all the facilities of the powerful 3D editor that comes with Unity.
However, creating new scenes in virtual environments might quickly become a time consuming procedure that requires experience in 3D graphics. This is even more evident when preparing photo-realistic objects, that requires the user to pay attention to a large number of aspects in order to reach a certain target appearance for the object.

In order to partially mitigate these issues, SAILenv comes with more than 65 objects that can be placed in any scene, plus some objects related to the structure of the sample scenes (walls, windows, etc.). See Fig.~\ref{fig:objects} for some examples of available objects. Most object meshes were originally taken from the AI2-THOR project, and strongly re-worked in order to improve the quality of their appearance, reaching a more advanced photo-realistic level.
In particular, we employed Physically Based Rendering (PBR), a state-of-the-art technique to define 3D mesh materials which correctly simulates the light on the mesh. Standard PBR shaders from the Unity built-in rendering pipeline were used to handle all the objects materials, and we manually tuned the ``Albedo, Metallic, Specular, Normal'' textures\footnote{See the full descriptions in the Unity documentation,  \url{https://docs.unity3d.com/Manual/StandardShaderMaterialParameters.html}.}.
Object models were properly edited for texturing, adding UV coordinates to them. To generate textures, we used the Substance\footnote{\url{https://www.substance3d.com/}} suite. 

In order to build the basic conditions for the global illumination of the environment, we used different HDRI (High Dynamic Range Imaging) skyboxes, a set of textures wrapped in cube-maps applied to the surrounding of a 3D scenes. HDRI skyboxes provide realistic environmental global illumination  and static reflections on all objects materials. Each one of them is based on real-world pictures. Static reflection probes\footnote{\url{https://docs.unity3d.com/Manual/class-ReflectionProbe.html}} and lights were added to generate additional objects reflections and to provide further illumination\footnote{At the time we are writing, the only limitation we report is that moving objects do not appear in reflections. We are working to improve this aspect without sacrificing too much the performance of the application.}. Finally, we applied a set of post-processing effects, using Unity Post-Processing Stack v2 package\footnote{\url{https://github.com/Unity-Technologies/PostProcessing/tree/v2}}, that we found to be a good trade-off between quality and impact in performance. 

\begin{figure*}[!t]
\centering
\subfloat[Room 01]{\includegraphics[width=0.2\textwidth]{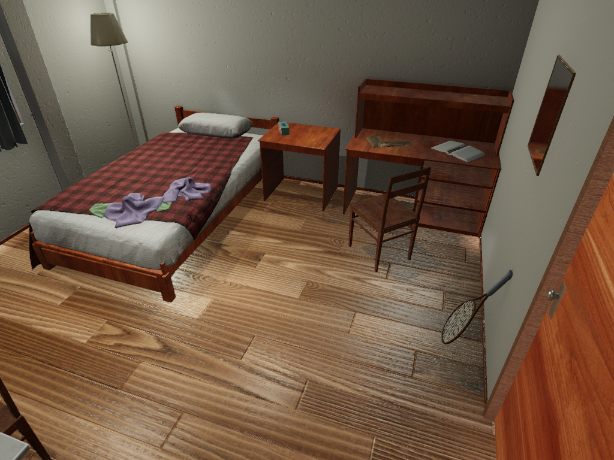}
\label{fig:room01}}
\hfil
\subfloat[Room 02]{\includegraphics[width=0.2\textwidth]{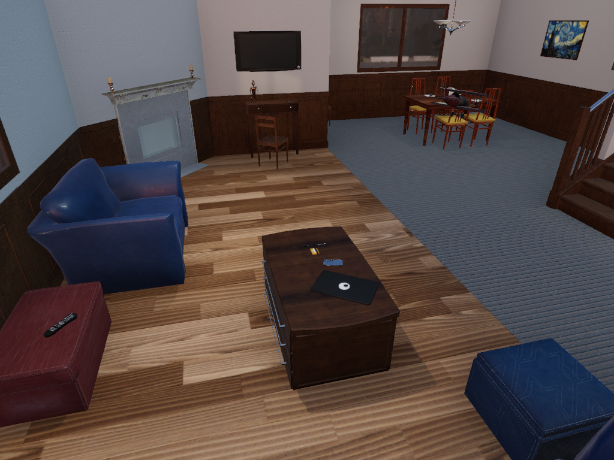}
\label{fig:room02}}
\hfil
\subfloat[Room 03]{\includegraphics[width=0.2\textwidth]{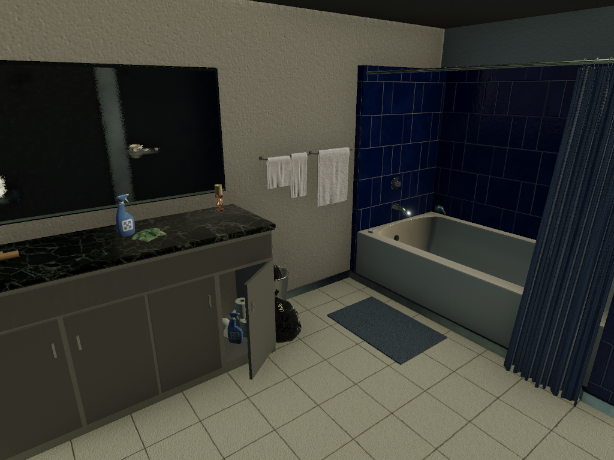}
\label{fig:room03}}
\hfil
\subfloat[Optical]{\includegraphics[width=0.2\textwidth]{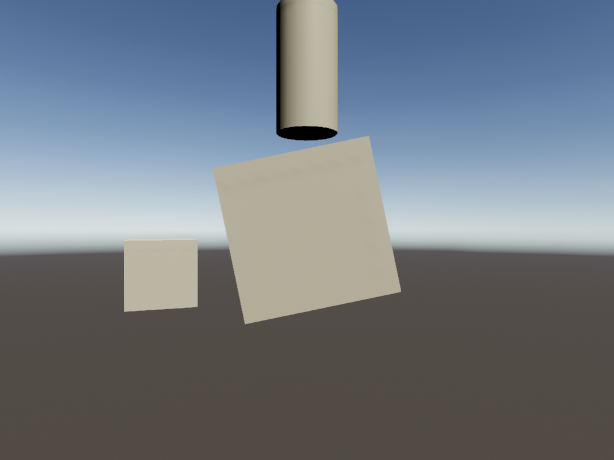}
\label{fig:optical_scene}}
\caption{Sample scenes in SAILenv}
\label{fig:sample_scenes}
\end{figure*}

SAILenv currently includes a ready-to-go Unity project with all the photo-realistic elements and 4 sample scenes based on them, meant to demonstrate the capabilities of the framework and to run some experiments in simple contexts, as we will describe in Section~\ref{sec:exp}. The user can either edit one of these scenes or create a new one either using the SAILenv objects or other 3D elements. The sample scenes are about different rooms, and they are based on a variety of objects, and they also include moving objects to evaluate motion-based algorithms. The agent has a predefined motion pattern such that it automatically moves around the scenes exploring the available areas (for the full descriptions on how to inject object and agent motion see Section~\ref{sec:moveobj} and Section~\ref{sec:mover}, respectively). 
In detail:
\begin{itemize}
    \item \textsc{Room 01}: Bedroom. Main objects: laptop, bed, desk, chairs and writing materials. See Fig.~\ref{fig:sample_scenes}a.
    \item \textsc{Room 02}: Sitting and dining areas. Main objects: chairs, couches, dining table, paintings. Object movement: a toy rusty plane flies around the room. See Fig.~\ref{fig:sample_scenes}b.
    \item \textsc{Room 03}: Bathroom. Main object: toilet, bathtub, cleaning supplies and hands towels. Object movement: many of the objects inside the scene will occasionally be pushed in a random direction, moving from their original position. See Fig.~\ref{fig:sample_scenes}c.
    \item \textsc{Optical}: It includes rotating cubes and a cylinder. This scene is not realistic and is meant to be used for debugging purposes (for example, to test the optical flow feature). See Fig.~\ref{fig:sample_scenes}d.
\end{itemize}

\subsection{Making Objects Move} 
\label{sec:moveobj}
Object movement is handled by the Unity physics engine, exploiting ``rigidbody'' components\footnote{\url{https://docs.unity3d.com/ScriptReference/Rigidbody.html}}. In particular, only those objects that have a rigidbody component attached to them can move, differently from the ones that are marked as ``full static''. Moving objects must define a mass value, which is not supposed to be completely realistic, but appropriate enough to generate a seemingly realistic behaviour. 

The concrete movement behaviour can be scripted within the Unity engine, using the C\# language. 
SAILenv includes three sample movement scripts that can be attached to multiple objects, and that were briefly mentioned when describing Room 02, Room 03, and Optical in Section~\ref{sec:photo}.
The first movement behaviour is formalized in the SAILenv script named ``poltergeist'', that randomly moves an object by applying both a force and a torque in random direction, at random time instants. 
The second movement behaviour is modeled by the ``wander plane'' script, that moves the rigidbody of an object along a configurable set of waypoints. Such waypoints are switched at random time intervals, and the resulting movement can be used, for instance, to emulate 
a flying airplane. There are a set of parameters that are configurable in the object inspector, to better customize this behaviour.
The last movement example included in SAILenv, ``rotate rigidbody'', is about the rotating elements of the Optical scene, that is indeed very simple and a useful reference for beginners.

\subsection{Handling Agent Movement}
\label{sec:mover}
Whenever an agent is created in the virtual environment, the Python client allows the user to define the position and the orientation of the agent. Of course, the user could develop his own Python routines to move the agent accordingly to some custom criteria. 
However, each SAILenv scene also includes a Unity object acting as track for the agents to follow. This object is called \textit{mover} within the Unity server, as shown in Fig.~\ref{fig:system_architecture}. This object is subject to the Unity physics engine, and all the agents that are active in the considered scene implicitly ``follows'' it, i.e., they inherit its position and orientation. This is implemented activating the ``follow rigidbody'' behaviour. 
In other words, the Python client asks a Unity agent to change its position, and this request is passed to the mover object that is the one that is actually moved to the new position. The outcome of this operation is that the agent will inherit the new position of the mover, thus the client will observe the agent moving. At a first glance, this organization might seem over-structured. However, SAILenv includes another facility that actually motivates such organization, that is the possibility of interacting with the virtual environment directly through the running instance of Unity in the server (thus not using the client at all).
In detail, whenever a SAILenv scene is created, a default agent, named \textit{debug agent}, can be added to the scene, whose properties are set in order to instruct him to mimic the mover object. Since both the debug agent and the client-manipulated agent follow the same mover, as shown in Fig.~\ref{fig:system_architecture}, the debug agent actually becomes a proxy of the client-manipulated agent (and vice-versa). 

SAILenv implements a ``rigidbody controller'' that allows the mover to be guided by means of the keyboard and mouse of the machine hosting the Unity server, in a 3D-shooter-like fashion. 
This is particularly useful for debug purposes, allowing the user to freely explore and see how the algorithm processing the client data will react.
The mover can also be controlled by an alternative policy the is implemented in the ``waypoint controller''.
In this case, the mover will follow a trajectory defined by a set of customizable ``waypoints'', that are placeholder objects with a given position/orientation.
For each pair of consecutive waypoints, the mover will linearly interpolate their positions/orientations and it will apply the interpolated transformation parameters to itself. 
The set of waypoints is configurable in each scene, as well as the total time to run across all the waypoints. When the last waypoint is reached, the mover will consider the first one and it will start another cycle. 
To switch between the two movement policies during the execution, the user can press the tab key on the Unity instance running on the server. 




\section{Using SAILenv}
\label{sec:using}
In this section we describe concrete examples of how to use the SAILenv platform, discussing both server and client-related operations. For an extended tutorial, documentation, and downloads, please refer to the project website, that is \url{http://sailab.diism.unisi.it/sailenv}. 

\subsection{Interfacing an Algorithm with SAILenv}
\begin{figure}[!ht]
    \centering
    \frame{\includegraphics[width=0.3\textwidth]{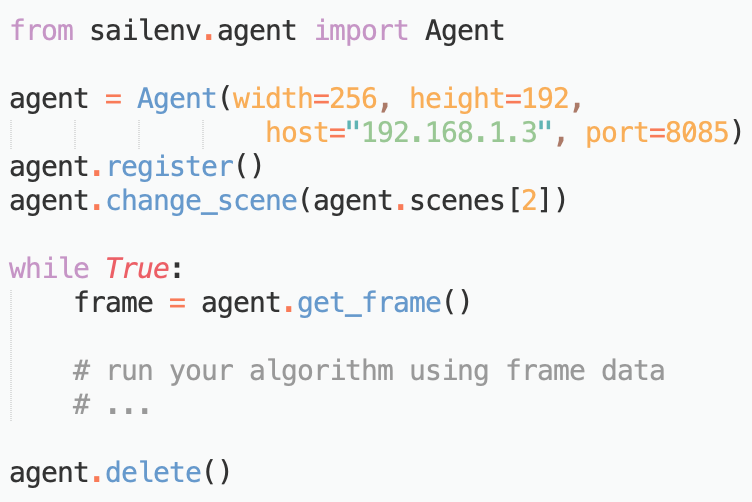}}
    \caption{Code that runs the Python client and get data to process.}
    \label{fig:code}
\end{figure}

The users aiming at interfacing an algorithm with SAILenv  have to take care of running the Unity server first, and then to import the package with the Python API in their own code, in order to generate a valid client.
The Unity editor allows the user either to directly run the current scene(s), thus activating the SAILenv Unity server as well, or to build the 3D scenes into valid executables for a target operative system, so that running an instance of the Unity server is trivial and it does not depend on the Unity editor. SAILenv comes with the scenes described in Section~\ref{sec:photo} that are pre-built and ready to be executed (we provide builds for the most common operative systems). Once the scene is running, the Unity server listens for connections on port 8085 (by default). The Python code needed to create a valid client and get data from the virtual environment is minimal, as shown in the snippet of Fig.~\ref{fig:code}.
The code initially creates and registers an agent on the Unity server. We assumed that the Unity Server is running the sample scenes of Section~\ref{sec:photo} on IP $192.168.1.3$. Notice that the resolution at which the agent will perceive the virtual environment is fully customizable when creating a new \texttt{\small Agent}. Also the views that are expected to be returned by the environment can be customized by using specific arguments.
The server runs a batch of scenes, that can be selected by calling the \texttt{\small agent.change\_scene(\textit{\footnotesize scene\_id})} method. The indices of the available scenes are in the \texttt{\small agent.scenes} array of the agent instance, so that in the code snippet we are selecting Room 02. This array is created after having registered the agent. The \texttt{\small agent.get\_frame()} method fetches the current views from the environment, that are returned in the dictionary \texttt{\small frame}. Such dictionary is composed by the following key-value pairs:
\begin{itemize}
    \item \textit{main}: $H \times W \times 3$ -- RGB view in OpenCV format (BGR).
    \item \textit{category}: $H \times W$ -- semantic labeling, in which each element at coordinates $(x,y)$ of the tensor is an integer containing the category ID associated to the object to which the pixel $(x,y)$ belongs. 
    \item \textit{object}: $H \times W \times 3$ -- instance labeling, that is a BGR image in which each pixel color is the unique identifier of the scene object to which the pixel belongs.
    \item \textit{flow}: $H \times W \times 2$ -- optical flow, composed of $v_x$, $v_y$ velocities of the flow.
    \item \textit{depth}: $H \times W \times 1$ -- the depth of each of the pixels of the agent camera, in $[0,255]$.
\end{itemize}
The final call to \texttt{\small agent.delete()} removes the agent from the server, releasing resources.

Going beyond the example of Fig.~\ref{fig:code}, another important property to mention is \texttt{\small agent.categories}, that is set when the agent registers or when the scene is changed, and that contains a dictionary that maps category numeric IDs to their respective names. By default, the agent will move in the virtual environments following the waypoints of the scene, as described in Section~\ref{sec:mover}. We can toggle this behaviour by calling \texttt{\small agent.toggle\_follow()}. Finally, we can change the position and orientation of the agent by \texttt{\small agent.set\_position(\textit{\footnotesize (x,y,z)})} and \texttt{\small agent.set\_rotation(\textit{\footnotesize (rx,ry,rz)})}, respectively. 

\subsection{Creating Scenes in SAILenv}
Once the SAILenv Unity project has been downloaded and opened in the Unity editor, the sample scenes are ready to be edited, or the user can trivially create a new scene from scratch. Adding each of the photo-realistic objects described in Section~\ref{sec:photo} to the current scene can be easily done by a drag-and-drop operation from the Unity project tab to the current scene or to the Unity hierarchy tab. Making an object move (Section~\ref{sec:moveobj}) requires to remove any full-static flags from the object and ensuring it has a rigidbody component attached to it.
In order to trigger the existing movement patterns described in Section~\ref{sec:moveobj}, it is necessary to add and/or turn on either the ``poltergeist'' or ``wander plane'' behaviour on an object with an attached rigidbody.
Of course, it is  possible to also make new behaviours by coding them directly, but this is outside the scope of this paper.

In the case of creation of new scenes from scratch, in order to make them compliant with the SAILenv framework, it is strongly suggested (even if not explicitly mandatory) to add a debug agent to it  (Fig.~\ref{fig:system_architecture}), while the client-related agent will be automatically created on the fly while registering the client. 
A very important element to add to the scene is the mover object (drag-and-drop), without which the agent (being it a debug or a client-related agent) cannot move. The debug agent, if present, must be connected to the mover, dragging the mover object into the Unity inspector of the debug agent (specifically into the ``target rigidbody'' field of the  ``follow rigidbody'' component). The mover already comes with the behaviours of Section~\ref{sec:mover}, i.e., 
``rigidbody controller'' and ``waypoint controller''. In order to design a trajectory along which the agent should move when the scene is running, several waypoints must be added to the scene. This can be easily done creating empty objects (placeholders) in positions and with rotations that we want the agent to visit, and then drag all these objects in the ``waypoints'' field of the ``waypoint controller'' behaviour inspector.


\section{Experimental evaluation}
\label{sec:exp}
We evaluated the concrete quality of SAILenv  photo-realism (Section~\ref{x}) and optical flow (Section~\ref{y}) exploiting state-of-the-art neural architectures.

\subsection{Photo-realism}
\label{x}
We prepared an experimental setting in which a state-of-the-art object detector (that also returns object masks), pretrained on real-world data, is interfaced with SAILenv. Focusing on the sample scenes of SAILEnv, we evaluated the capability of such model to correctly recognize  the object categories it is aware of, thus indirectly measuring how strongly the appearance of SAILenv objects resembles the one of the corresponding real-world objects.
In particular, we exploited a Mask R-CNN model based on the popular ResNet-50 backbone \cite{matterport_maskrcnn_2017}, pretrained on the COCO-train2017 data \cite{lin2014microsoft}. 
We focused on a subset of the categories of the COCO data, in particular in $14$ classes that are shared with SAILenv objects.
For each SAILenv object, we collected 5 frames using the SAILenv client, in different viewing conditions. The list selected categories is reported in the first column of Table~\ref{tab:segment}.

For each class, we measured the average Intersection over Union (IoU) between the pixel-wise predictions obtained through the Mask R-CNN mask branch and the ground truth   categories returned by SAILenv. Results are reported in second column of Table~\ref{tab:segment}.
Mask R-CNN is able to identify in a very robust way a large portion of the objects, despite the different viewing condition and the virtual setting. The produced masks mostly overlap the ground truths returned by SAILenv, with some mild exceptions that are due to the labeling criteria that are followed in the COCO training data, slightly different from the ones in SAILenv. 
For example, in tennis racket and potted plant, the masks predicted by Mask R-CNN tend to occupy all the area of the object, differently from the highly detailed pixel-wise labels produced by SAILenv, in which the spaces among the leaves of the plant, or within the net of the racket, are not marked with the object label. In the case of dining table, chair and, more generally, when there are some occlusions, we observed a related behaviour. The book in SAILenv is different from the ones used in the training data of COCO, while spoon and fork are frequently in different viewing conditions with respect to the ones in the COCO data.
To further investigate these intuitions, we also computed a measure that took care of evaluating how strongly the bounding box of the predicted regions matched the ones of the ground truth, leading to the third column of Table~\ref{tab:segment} (bounding box IoU). 
In this case, we observed that the prediction quality increased, on average, overcoming some of the aforementioned issues (see, e.g., potted plant, tennis racket). 
We clearly see large standard deviations around some other critical categories (spoon, fork, dining table, chair), confirming that the recognition succeeded in some of the viewing conditions and failed in others, as previously claimed.
\begin{table}[]
    \centering
    \caption{Mean and standard deviation of the predictions of the Mask R-CNN model (pretrained on COCO2017 dataset) on a dataset obtained from the SAILenv sample scenes. Two measures are considered: Pixel-wise IoU and Bounding Box IoU (see the paper text for details). }
    \begin{tabular}{c|c|c}
        \toprule    
        \textsc{Category}  & \textsc{Pixel-wise IoU} &  \textsc{Bounding Box IoU} \\
        \midrule
        bed & 0.7830 $\pm$ 0.0879 & 0.8201 $\pm$ 0.0894 \\      
        book & 0.3347 $\pm$ 0.2749 & 0.3506 $\pm$ 0.2870 \\
        chair & 0.6235 $\pm$ 0.0566 & 0.5557 $\pm$ 0.4162 \\
        couch & 0.8742 $\pm$ 0.0533 & 0.9121 $\pm$ 0.0561 \\
        dining table & 0.6891 $\pm$ 0.0398 & 0.4553 $\pm$ 0.4096 \\
        fork & 0.4599 $\pm$ 0.1274 & 0.4800 $\pm$ 0.4294 \\
        laptop & 0.9551 $\pm$ 0.0098 & 0.9476 $\pm$ 0.0207 \\
        airplane & 0.7193 $\pm$ 0.0314 & 0.7865 $\pm$ 0.1005 \\
        potted plant & 0.6106 $\pm$ 0.0499 & 0.8894 $\pm$ 0.0656 \\
        remote & 0.8980 $\pm$ 0.0400 & 0.9534 $\pm$ 0.0127 \\
        spoon & 0.4036 $\pm$ 0.1984 & 0.3787 $\pm$ 0.3611 \\
        tennis racket & 0.5120 $\pm$ 0.0475 & 0.9548 $\pm$ 0.0127 \\
        toilet & 0.9274 $\pm$ 0.0178 & 0.9623 $\pm$ 0.0201 \\
        tv & 0.9641 $\pm$ 0.0171 & 0.9673 $\pm$ 0.0135 \\
        \bottomrule
    \end{tabular}
    
    \label{tab:segment}
\end{table}

\subsection{Optical Flow Computation}
\label{y}
In Section \ref{sec:flow} we described how SAILenv generates a dense optical flow that is not an estimation performed observing pairs of frames, but it is about the real motion information coming from the 3D environment.
Of course, this leads to the most accurate motion estimation one could have. However, we are left with the open question on the computational burden with respect to competing algorithms, such as the popular OpenCV implementation of the Farneback algorithm \cite{farneback2003two}, and one of the fastest models based on convolutional neural networks, that is  LiteFlowNet \cite{flownetlite}.

\begin{figure}[H]
\begin{center}
\includegraphics[width=0.8\linewidth]{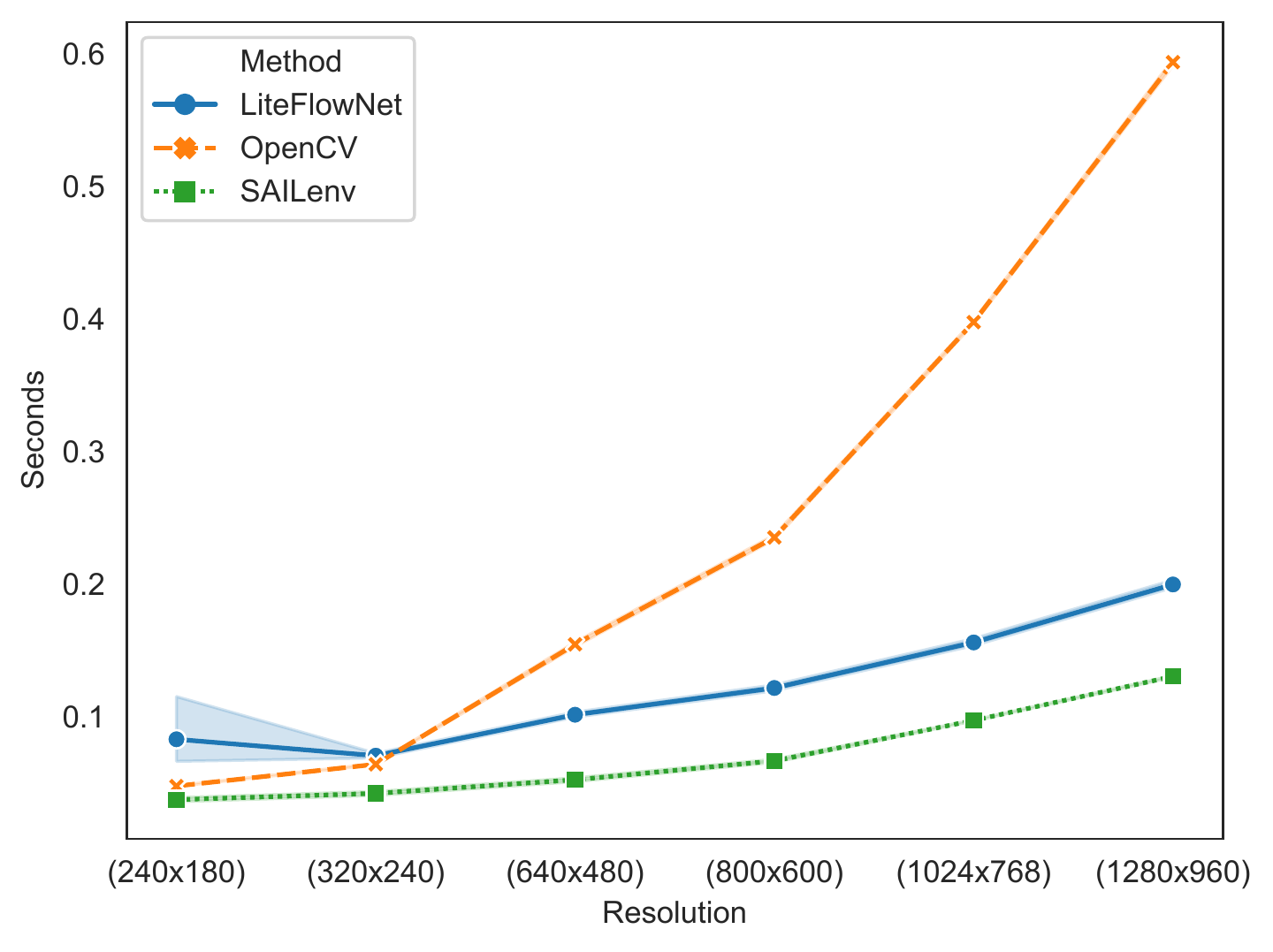}
\caption{Average time (seconds) needed to compute the optical flow associated to a frame sampled from the SAILenv scenes. We compare the SAILenv performances with an OpenCV-based implementation of the Farneback algorithm and the neural model LiteFlowNet\cite{flownetlite}.} 
\label{fig:flow_comparison}
\end{center}
\end{figure}

The first competitor exploits the OpenCV tools to speed up the computation, which, in the default Python distribution, is performed using the CPU. 
Differently, in the case of LiteFlowNet, we considered a PyTorch implementation\footnote{\url{https://github.com/sniklaus/pytorch-liteflownet}} which leverages GPU-based computations (CUDA).
For each compared method, we measured the time needed to produce the flow at six different resolutions, reported in the x-axis of Fig.~\ref{fig:flow_comparison}. On the y-axis we reported the average time over 100 sampled frames, with $95\%$ confidence intervals. 


In our experimental setting, when considering the OpenCV implementation and LiteFlowNet, SAILenv returns only the RGB frame from the virtual environment, turning off all the internal optical flow computation facilities. Then, the optical flow is computed using one of the competitors, exploiting the current frame and the one returned at the previous time step. The time needed to transfer data to/from the GPU was subtracted from our measurements.
We used a Windows desktop machine equipped with an Intel Core i9 9900K, $3.60$ GHz, 64 GB of RAM and an NVIDIA GTX 1080 GPU with 8 GB of VRAM. We also performed experiments on two other machines with different configurations, obtaining results with an analogous trend. 
SAILenv outperforms the competitors, sometimes reaching almost real-time performances, thanks to the direct access to the  physics engine of Unity that is strongly optimized and GPU-based. The gap with the competitors is pretty evident especially at high resolutions.

\section{Conclusions and Future Work}
\label{sec:conclusions}
We presented \textit{SAILenv}, the Siena Artificial Intelligence Lab environment, a software platform that makes it easy to create, run, and get data from realistic 3D virtual environments, on which visual recognition or other vision-related algorithms can be efficiently evaluated. SAILenv is based on a server that exploits the Unity engine, performing real-time rendering. SAILenv comes with a Python client and a collection of ready-to-use photorealistic 3D elements that can be easily assembled to create new scenes within the Unity editor, whose quality has been assessed by an experimentation based on a state-of-the art object detector. To the best of our knowledge, SAILenv is the first platform that allows researchers to have access  to motion information inherited by the 3D engine, thus extremely accurate and efficiently computed, as we evaluated in a comparison with other popular optical flow algorithms. Due to its simplicity, we believe that SAILenv will provide several researches an efficient entry point to 3D virtual environments. Future work will include the extension to multi-agent systems and new photorealistic objects and scenes.

\bibliographystyle{IEEEtran}
\bibliography{biblio}
\end{document}